\documentclass[runningheads,a4paper]{llncs}

\usepackage{amssymb}
\usepackage{graphicx}
\usepackage{algorithm} %ctan.org\pkg\algorithms
\usepackage{algpseudocode}
\usepackage{adjustbox}
\usepackage{amssymb,amsfonts,amsmath}
\usepackage{booktabs}
\newcommand{\ra}[1]{\renewcommand{\arraystretch}{#1}}

\usepackage{url}
\urldef{\mailsa}\path|{wen.shen, lopes}@uci.edu|
\newcommand{\keywords}[1]{\par\addvspace\baselineskip
\noindent\keywordname\enspace\ignorespaces#1}

\algrenewcommand\algorithmicrequire{\textbf{Precondition:}}

\begin{document}

\mainmatter  % start of an individual contribution

\title{Managing Autonomous Mobility on Demand Systems for Better Passenger Experience}

\titlerunning{Managing Autonomous Mobility on Demand Systems}

\author{Wen Shen
\and Cristina Lopes}
\authorrunning{W. Shen et al.}

\institute{Department of Informatics\\
University of California, Irvine, CA92697, USA\\
\mailsa\\}


\toctitle{Lecture Notes in Computer Science}
\tocauthor{Authors' Instructions}
\maketitle

\begin{abstract}

 Autonomous mobility on demand systems, though still in their infancy, have very promising prospects in providing urban population with sustainable and safe personal mobility in the near future. While much research has been conducted on both autonomous vehicles and mobility on demand systems, to the best of our knowledge, this is the first work that shows how to manage autonomous mobility on demand systems for better passenger  experience. We introduce the Expand and Target algorithm which can be easily integrated with three different scheduling strategies for dispatching autonomous vehicles. We implement an agent-based simulation platform and empirically evaluate the proposed approaches with the New York City taxi data. Experimental results demonstrate that the algorithm significantly improve passengers' experience by reducing the average passenger waiting time by up to $29.82\%$ and increasing the trip success rate by up to $7.65\%$.
\end{abstract}

\section{Introduction}
\label{introduction}
As urbanization accelerates and city population continues to grow, more traffic is generated due to the increasing demand for personal mobility as well as the upswing of private car ownership~\cite{downs2005still}. This  results in many severe problems such as  traffic congestion, air pollution and limited public space available for the construction of parking areas and roads~\cite{beirao2007understanding}. To solve these problems, it is in urgent need of building a transportation system that not only satisfies people's mobility demand but is also more sustainable, efficient and reliable. Although autonomous mobility on demand (AMOD) systems, which, due to some technical, economical and governmental obstacles left to be overcome,  are still in their infancy, they hold very promising prospectus in meeting the need. This is because AMOD systems have great potential to provide a safer, near instantly available solution for personal mobility. Once such services have become highly available and affordable, the collective transport solutions will eventually transcend the traditional private ownership, which will significantly reduce the traffic congestion, pollution and land use for parking purpose through system-wide optimization and coordination. Besides, autonomous vehicles can also avoid accidents caused by human errors, making them safer than conventional cars driven by human drivers~\cite{lozano2012autonomous}.

In recent years, much research has been conducted on autonomous vehicles. However, most of the research focuses on the control of a single autonomous vehicle to perform various tasks including picking up passengers and parking. While interesting and important,  it leaves much other territory uninvestigated, especially methodologies on managing systems of autonomous vehicles for personal mobility. To bridge the gap and address the transportation problems brought by city expansion and population growth, we study how to dispatch AMOD systems to improve passengers' experience. In doing so, we introduce the \textit{Expand and Target} (EAT) algorithm. We then conduct agent-based simulation using the AMOD simulation platform based on the \textit{MobilityTestbed}~\cite{vcerticky2014agent}\cite{vcerticky2015analyzing} and the New York City taxi data~\cite{dan2014newyork}. 

The rest of the paper is organized as follows: Section~\ref{sec:background} briefly discusses the background and related work on autonomous vehicles, mobility on demand(MOD) systems,  and AMOD systems; Section~\ref{sec:managing} introduces three scheduling strategies and the \textit{Expand and Target} algorithm for managing  AMOD systems; Section~\ref{sec:experiment} talks about the experiments for evaluating the proposed dispatching approaches for AMOD systems. Section~\ref{sec:conclusion} concludes this paper and presents potential directions for further investigation.

\section{Related Work}
\label{sec:background}

\subsection{Autonomous Vehicles}
An autonomous or automated vehicle is a vehicle capable of fulfilling transport tasks such as motion and braking without the control of a human driver~\cite{lozano2012autonomous}. The development of autonomous vehicles relies on advances in computer vision, sensing technology, wireless communication, navigation, robotic control as well as technologies in automobile manufacturing~\cite{anderson2014autonomous}. Significant progresses have been achieved in these fields over the past several decades. For example, LIDAR and vision-based detection techniques (e.g., stereo vision) are extensively studied in pedestrian, lane and vehicle detection~\cite{premebida2009lidar}\cite{huang2009finding}\cite{moghadam2008improving}. Vehicular communication systems make it possible for individual vehicles to share information (e.g., traffic congestion, road information) from other vehicles in the vicinity, which can potentially improve the operational safety of autonomous vehicles~\cite{papadimitratos2009vehicular}.

The shift from conventional cars to autonomous vehicles can substantially reduce traffic accidents caused by human errors, given the fact that the operation of autonomous vehicles does not involve human intervention~\cite{anderson2014autonomous}. It also increases mobility for people who are unable or unwilling to drive themselves~\cite{anderson2014autonomous}. 

As technology advances, many prototypes of autonomous cars have been designed and developed. Some examples of these include Google driver-less car~\cite{markoff2010google}, Vislab's BRAiVE~\cite{broggi2013extensive},  and BMW's M235i~\cite{lavrinc2014bmw}, just to name a few. These cars are also successfully tested in various real-world traffic environments, making it possible and desirable to be integrated with MOD systems.

\subsection{Mobility on Demand Systems}
To reduce private car ownership while also meeting the need for personal urban mobility, Mitchell et al.  introduces MOD systems~\cite{mitchell2010reinventing}\cite{chong2013autonomy}. The original purpose of MOD systems is to complement mass transportation systems such as subways and buses, providing commuters with the mobility for the first mile and the last mile. A notable prototype is the CityCar~\cite{mitchell2007intelligent}. 

The vehicles in current MOD systems are mainly light electric vehicles across the main stations around the city , guided  by human drivers or guideways,  which limits the scope of mobility and the development of the MOD systems. If autonomous vehicles were integrated into the MOD systems, the AMOD systems would transform mobility and revolutionize the transportation industry by offering  more flexible and more convenient solutions for urban personal mobility.

\subsection{Autonomous Mobility on Demand Systems}
Due to lack of infrastructure, little research has been done on autonomous MOD systems.  Among the very few studies, Spieser et.al.~\cite{spieser2014toward} provides analytical guidelines for autonomous MOD systems using Singapore taxi data. The results show that AMOD systems could meet the mobility need of the same population with only 1/3 of the taxis that are current in operation. Zhang et al.~\cite{zhang2014control} presents a similar analysis using a queueing-theoretical model with trip demand data in Manhattan, New York City. While interesting and innovative, the research leaves a number of issues unexplored. For instance, how to efficiently manage autonomous vehicles to enhance passengers' experience? 

Although numerous research addresses taxi dispatching problems using various methodologies including techniques in multi-agent systems~\cite{alshamsi2009multiagent}\cite{seow2010collaborative}\cite{glaschenko2009multi}\cite{agussurja2012toward}\cite{gan2015optimizing}, no or at least little research addresses methodologies on dispatching AMOD systems for better user experience. Unlike conventional taxi dispatching systems where it is often difficult to coordinate because of human factors (e.g., the drivers may not follow the dispatcher's instructions carefully), the AMOD systems make it possible to implement a much higher level of autonomy through delicate, system-wide coordination. Moreover, it is challenging to achieve near instantly availability using conventional approaches. We try to bridge this gap by introducing a new algorithm-the \textit{Expand and Target} algorithm to effectively manage AMOD systems.

\section{Managing Autonomous Mobility on Demand Systems}
\label{sec:managing}
In this section, we first discuss three scheduling strategies that are either commonly used in the taxi dispatching field or frequently studied in literature: the No-Scheduling Strategy (NSS), the Static-Scheduling Strategy (SSS), and the Online-Scheduling Strategy(OSS)~\cite{maciejewski2013simulation}. 

For effective management of AMOD systems, we introduce the \textit{Expand and Target} (EAT) algorithm. This algorithm enables managing authorities of AMOD systems to automatically and effectively dispatch autonomous vehicles and meanwhile update adjacency schedule as well for better passenger experience: first, it increases the possibility of finding global optimal solutions; second, it reduces computation time by avoiding looping all the possible vehicles; third, it connects isolated areas with other dispatching areas and updates the adjacency schedule automatically, which increases the dispatch success rate.

We then discuss methodologies on integrating the scheduling strategies with the EAT algorithm to improve passengers' experience, especially to reduce the average passenger waiting time and increase the trip success rate.
\subsection{Scheduling Strategies}
\label{scheduling}
\subsubsection{No-Scheduling Strategy}
In this strategy, the dispatcher assigns the nearest idle taxi to an incoming call. If such a taxi can not be found, then the call will be rejected. The dispatcher does not update the dispatching schedule. That is why we call it the No-Scheduling Strategy or NSS. The NSS is the most commonly used strategy in taxi dispatching applications~\cite{seow2010collaborative}.

\subsubsection{Static-Scheduling Strategy}
In SSS, the dispatcher keeps a schedule of the dispatching process and updates it by appending an incoming call to the list of processed requests.  When a new call comes, the dispatcher searches for the nearest taxi from all the vehicles (both idle and busy). It then estimates the time from current position to the pickup location based on current traffic condition. For a idle taxi, the dispatcher simply computes a trip plan directly from the taxi's current location to the pickup location. While for a taxi that is busy, the dispatcher then calculates the remaining time needed for the taxi to complete the current trip and then calculate a trip plan from the end point of the trip to the pickup location. Then the dispatcher estimates the total time needed for this taxi to arrive at the pickup location. In this way, the dispatcher selects a taxi that can reach the passenger to be picked up in the quickest time. The dispatcher does update the schedule of the dispatching process, but it never reschedules or reassigns the requests. This strategy broadens the choice of taxis, but it scales poorly due to one-time scheduling.

\subsubsection{Online-Scheduling Strategy}
The OSS is similar to the static approach except that in the online strategy the schedules are always re-computed in response to traffic variations such as delays or speedups. It is more cost efficient than SSS but requires more computational power.
\subsection{The Expand and Target Algorithm}
 We formally introduce the \textit{Expand and Target} algorithm (see Algorithm \ref{etalgorithm}).  The basic idea of the algorithm is described as follows: 
\begin{itemize}
\item When the dispatcher receives a call $c$, it first identifies the neighborhood $a_{c}$ that $c $ originates from.
\item If area $a_{c}$ has no adjacent areas, it then searches for the nearest available vehicle in area $a_{c}$: if found, it assigns the vehicle to the call, and returns the assignment;  if not, it searches for the nearest available vehicle in all dispatching areas: if found,  it adds the vehicle's area $a_{i}$ as a neighbor of area $a_{c}$ (by doing so, it removes the isolated dispatching area and updates the adjacency schedule), assigns the vehicle to the call, and  returns the assignment; if not found, it rejects the call. Meanwhile, it updates the schedule.
\item If area $a_{c}$ has other adjacent areas $\tilde{B_{a_{c}}}$ where the service is available, then we define the dispatching area for call $c$ as $B_{a_{c}}$. Instead of searching for the nearest available vehicle in $a_c$, it first expands the dispatching area for $c$:  $B_{a_{c}} \gets \tilde{B_{a_{c}}} \cup a_{c}$. We call this process as \textit{expand}. The expansion is necessary because it diminishes the possibility of finding local minimum: For example, when a call is from the border of several areas, it is not sufficient to decide whether a nearest available vehicle is in the current dispatching area or from other areas without considering all the neighborhoods. 
\item  After expansion,  then \textit{target}: it searches for the nearest available vehicle in $B_{a_{c}}$: if found,  it assigns the vehicle to the call, returns the assignment and terminates; if not found, then continues to expand the dispatching area $B_{a_{c}}$ using the previous strategy. 
\end{itemize}

\begin{algorithm}
\caption{\textit{Expand and Target}}\label{etalgorithm}
\begin{algorithmic}[1]
 \Require{$c$ -an incoming call,  $V$ - autonomous vehicles in operation, and $A$-dispatching adjacency schedule}    
 \Statex
\Procedure{Expand and Target}{$c, V, A$}
  
   \State $\textrm {identify } c's \textrm{ dispatching area } a_{c} \in A$
   
   \If{$\textrm{area } a_{c} \textrm{ has adjacent areas (immediate neighbors) } \tilde{B_{a_{c}}}  \subseteq A$}
   \State $//\textrm{begin to \textit{expand: }}$
   \State $ B_{a_{c}} \gets a_{c} \cup \tilde{B_{a_{c}}}$
   \State $// \textrm{begin to \textit{target: }} $
   \If{$\textrm{there are available vehicles in area } B_{a_{c}}$}
   \State $\textrm {in area } B_{a_{c}} \textrm{, search for the nearest available vehicle } v \in V$   
    \State \Return{\textrm{assignment pair } (v, c) }

   \Else
   \While{$a \textrm{ in } B_{a_{c}}$}
   \State{$\textrm{continue to \textit{expand} within A}$}
   \EndWhile
   \State \Return{\textrm{reject the call } c}
   \EndIf
   \Else
   	\If{$\textrm{there are available vehicles in area } c$}
   		\State $\textrm {in area } c \textrm{, search for the nearest vehicle } v \in V$
   		 \State \Return{\textrm{assignment pair } (v, c) }
   	\Else
   		\If{$\textrm{there are available vehicles in area } A$}
   		\State $\textrm {in area } A \textrm{, search for the nearest vehicle } v \in V$  
   		   \State $// \textrm{begin to \textit{update: }} $ 
   		\State $\textrm {set the dispatching area } v  \textrm{ as a neighbor of  area } a_{c}$	
   		 \State \Return{\textrm{assignment pair } (v, c) }
   		\Else
   		\State $\textrm{reject the call } c$
   		\EndIf
   	\EndIf
   \EndIf
\EndProcedure
\end{algorithmic}
\end{algorithm}

The \textit{Expand and Target} algorithm dynamically expands the search space and targets the autonomous vehicles, in which the expansion and targeting can be viewed as a multi-agent, self-adaptive process. The  algorithm assumes that the dispatching system processes passengers' requests in a First-Come, First-Served (FCFS) manner. It is also suitable for dispatching systems using other policies such as batch processing. However, modification of the algorithm is necessary for such applications (though we do not explore it in this work). 
\subsection{Integration}
\label{sec:integration}
To improve the performance of the dispatching system, we combine the three scheduling strategies discussed in section~\ref{scheduling} and the EAT algorithm, resulting in three integrated approaches:  NSS-EAT, SSS-EAT, and OSS-EAT. The incorporation is straightforward: when applying the EAT algorithm for dispatching autonomous vehicles, the dispatcher computes the nearest available vehicles using the three different scheduling strategies respectively. There is a commonly used searching method for taxi dispatching: when a call comes, the dispatcher first searches for a nearest available taxi in the current dispatching area; if such a taxi not found, then searches for an available taxi in an adjacent area. If the taxi does not have adjacent areas, the dispatcher then rejects the call. We also integrate this method with the three scheduling strategies as three control groups: NSS, SSS, and OSS.

\subsection{The Autonomous Mobility on Demand Simulation Platform}
We implement the simulation platform for AMOD systems on top of the \textit{MobilityTestbed}~\cite{vcerticky2014agent}\cite{vcerticky2015analyzing}. The \textit{MobilityTestbed} is an agent-based,  open-source simulation framework (written in Java) tailored for modeling and simulating on-demand transport systems. It consists of three layers: the \textit{AgentPolis} simulator~\cite{jakob2012agentpolis}, the Testbed Core and the mechanism implementation~\cite{vcerticky2014agent}\cite{vcerticky2015analyzing}. The first two layers provide the Testbed API while the third layer enables  users to implement their own mechanisms or strategies for various on-demand transport systems. The mechanism implementation includes three agent logic blocks: the driver agent logic, the dispatcher agent logic and the passenger agent logic. In AMOD systems, no drivers are needed, so we replace the driver logic with the autonomous vehicle agent logic. The \textit{MobilityTestbed} does not work properly with large datasets (e.g., a trip demand file larger than 1GB) due to poor memory management schemes. Thus, We modify the Testbed Core to make the platform be able to support large datasets.

This testbed also contains other components such as experiment management, benchmark importer, and visualization and reporting. However, the original experiment management component can only produce overall statistics, whereas in real-world scenarios it is important and necessary to log the operational data periodically for future use. The benchmark importer component provides benchmark data for studying on-demand transport systems, but it is customized for traditional on-demand transport service and only deal with small datasets. The visualization and reporting component, using Google Earth to visualize daily mobility pattern of passengers and vehicles, does not work properly with large-scale datasets. For these reasons, we discard all the three components. To meet the needs of the experiment, we implement the data logger component which enables the platform to efficiently log the operational data periodically. 

\begin{figure}
\centering
\caption{An overview of the Autonomous Mobility on Demand Simulation Platform based on \textit{MobilityTestbed}.}
\includegraphics[width=\textwidth]{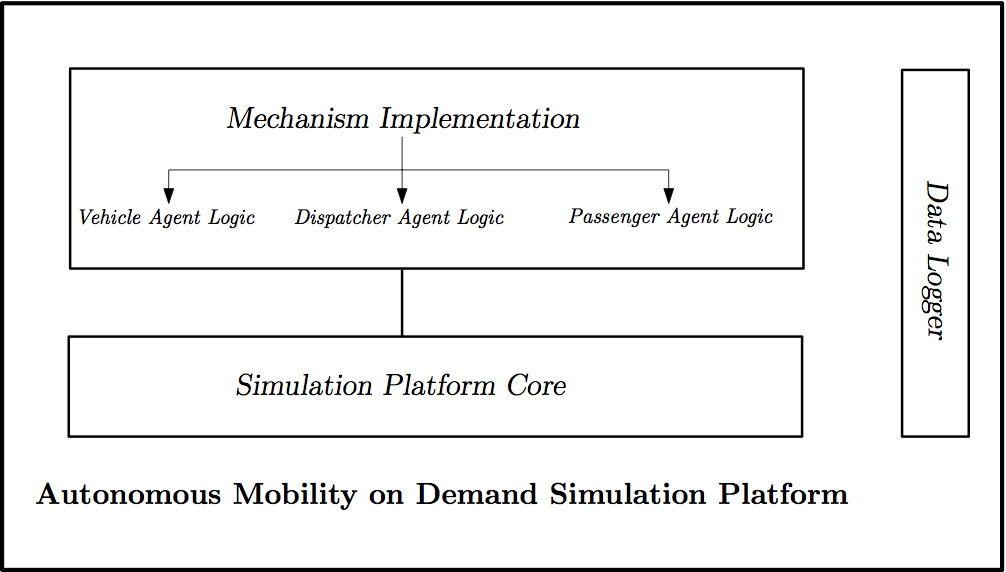}
\label{fig:overviewamodsp}
\end{figure}

An overview of the AMOD platform based on the \textit{MobilityTestbed} is shown in Fig.~\ref{fig:overviewamodsp}. The simulation platform is composed of two layers- the mechanism implementation layer and the simulation platform core layer, and an extension-the data logger. The mechanism implementation includes three logic blocks: the vehicle agent logic, the dispatcher agent logic, and the passenger agent logic. The simulation platform core is built on the basis of the \textit{MobilityTestbed} which consists of the testbed core and the \textit{AgentPolis} platform.

\section{Experimental Analysis}
\label{sec:experiment}
To empirically evaluate the performance of the \textit{Expand and Target} algorithm on managing AMOD systems, we implement six different dispatching approaches using the AMOD simulation platform on the basis of the \textit{MobilityTestbed} and OpenStreetMap~\cite{haklay2008openstreetmap}.  We then perform experiments using the 2013 New York City taxi data~\cite{dan2014newyork} and analyze the experimental results. 

\subsection{Evaluation Metrics}
To evaluate the performance of the dispatching system from the passengers' perspective, we select the following two metrics: the Average Passenger Waiting Time ($T_{APW}$) and the Trip Success Rate ($R _{TS}$). This is because they are considered as the two most important indicators of passenger satisfaction (quality of service)~\cite{maciejewski2013simulation}. 

The average passenger waiting time is formulated as the following equation:
\begin{equation}
\label{eq:tapw}
T_{APW} = \frac{\sum_{i \in N}  (T_{i}^{p}-T_{i}^{r})}{\tau}  ,
\end{equation}
where $T_{i}^{p}$ and $T_{i}^{r}$ are the pickup time  and the request time of call $i$, respectively. $N$ is the set of calls and $\tau$ is the number of calls in set $N$.

The trip success rate is defined as below:
\begin{equation}
\label{eq:rts}
R_{TS} = \frac{n_{s}}{n}  ,
\end{equation}
where $n_{s}$ is the number of successful trips and $n$ is the number of calls.

\subsection{The Datasets}
We choose the 2013 New York City Taxi Data~\cite{dan2014newyork} to generate the trip demand.  This raw data takes up for about 21.23GB in CSV format, in which each row represents a single taxi trip (including demand information). Table~\ref{table:datasample} shows a small sample of data which we use in the experiments. In order to compute the average passenger waiting time, the request time of the calls is needed. However, there is no such information available in this dataset. To solve this problem, we use the actual pickup time as the request time. We compute the pickup time and dropoff time using the $A^{\ast}$ routing algorithm~\cite{delling2009engineering}. Once a vehicle has finished a trip, it immediately become idle unless a new assignment comes. In the experiment, a passenger is assigned with a patience value, which indicates the longest time that the passenger would like to wait for a car before he/she cancels the request. We randomly generate the patience value for each passenger from 60 seconds to 3600 seconds.

Due to failure or faults of data collection devices, the NYC Taxi Data contains a large number of errors such as impossible GPS coordinates (e.g., (0,0)), times, or distances.  To remove these errors, we conduct data preprocessing after which there are 124,178,987 valid trips left in total. The fleet consists of 12,216 autonomous vehicles with the same configurations such as load capacity and speed capability.
\begin{table}
\caption{A small subset of the data used in trip demand generation} % title of Table
\centering % used for centering table
\begin{adjustbox}{width=1\textwidth}
\begin{tabular}{c c c c c c c c} % centered columns (7 columns)
\hline\hline %inserts double horizontal lines
medallion & pickup time & dropoff time& passenger count & pickup log & pickup lat& dropoff log&dropoff lat \\ [0.5ex] % inserts table
%heading
\hline % inserts single horizontal line
2013002932&2013-01-02 23:43:00&2013-01-02 23:49:00&4&-73.946922&40.682198&-73.92067&40.685219\\\\

2013009193&2013-01-02 23:43:00&2013-01-02 23:54:00&2&-74.004379&40.747887&-73.983376&40.766918\\\\

2013007140&2013-01-02 23:46:00&2013-01-02 23:55:00&1&-73.869743&40.772369&-73.907898&40.767262\\\\

2013008400&2013-01-02 23:50:12&2013-01-02 23:56:41&3&-73.984756&40.768322&-73.983276&40.757259\\
\hline %inserts single line
\end{tabular}
\end{adjustbox}
\label{table:datasample} % is used to refer this table in the text
\end{table}

\begin{figure}
\centering
\caption{Initial adjacency schedule from  Pediacities New York City Neighborhoods GeoJson data shown on OpenStreetMap}
\includegraphics[width=0.7\textwidth]{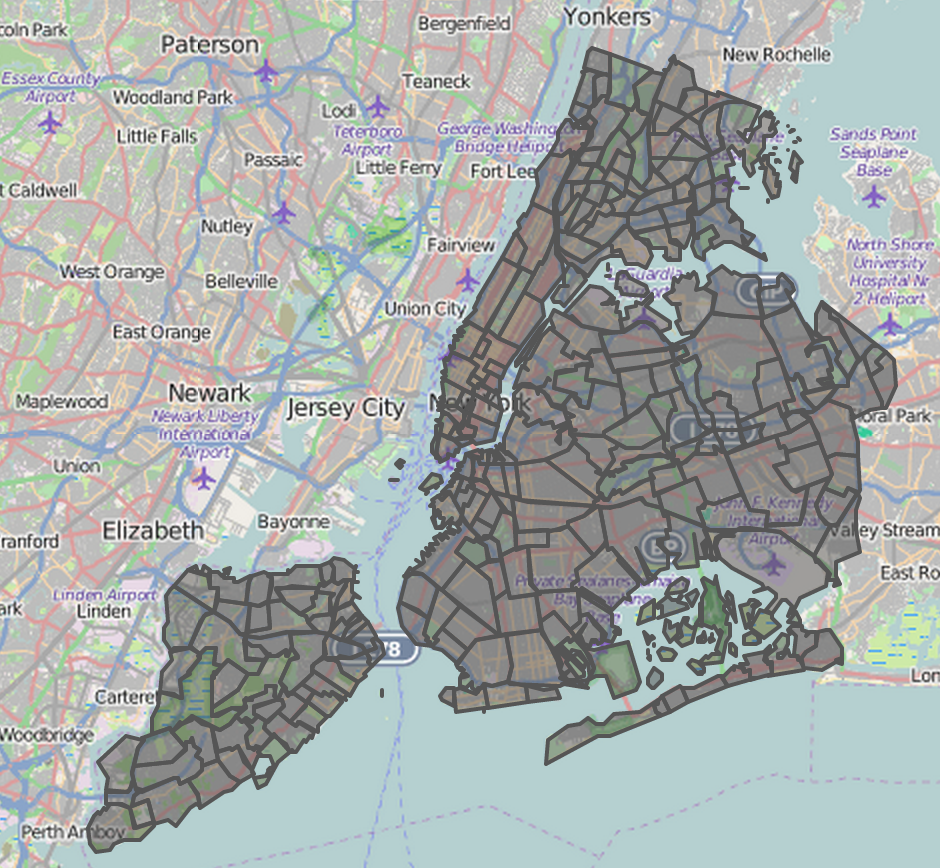}
\label{fig:initialadj}
\end{figure}

\begin{figure}
\centering
\caption{Final adjacency schedule using the \textit{Expand and Target} algorithm shown on OpenStreetMap}
\includegraphics[width=0.7\textwidth]{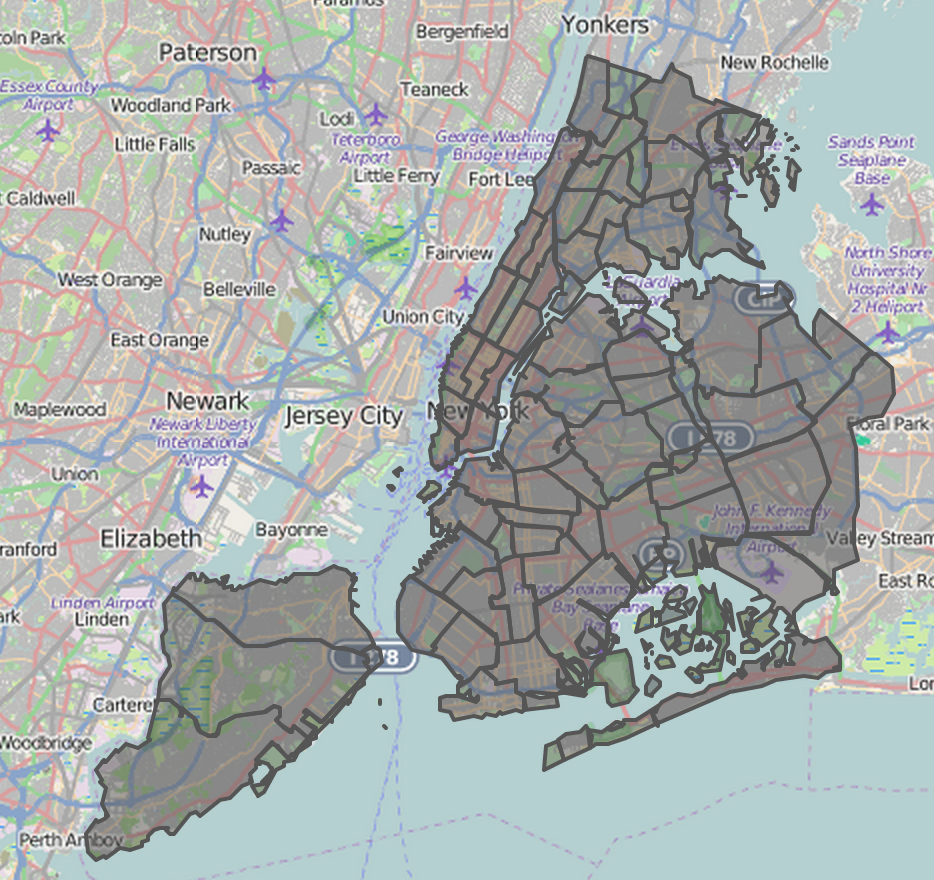}
\label{fig:finaladj}
\end{figure}

The initial dispatching adjacency (see~Fig.~\ref{fig:initialadj}) is generated from the  Pediacities New York City Neighborhoods GeoJSON data~\cite{pediacities2015nyc}, in which there are 310 neighborhood boundaries in total. Fig.~\ref{fig:finaladj} shows the final adjacency schedule (71 neighborhood boundaries) using the \textit{Expand and Target} algorithm with New York taxi data.

As for the navigation map, we use the OpenStreetMap data for New York City obtained from MAPZEN metro extracts~\cite{mapzen2015nyc} on April 9th, 2015. The size of the OSM XML data file is about 2.19GB.

\subsection{Experimental Settings}
In the experiment, we implement the three integrated approaches described in section~\ref{sec:managing}: the NSS-EAT, the SSS-EAT, and the OSS-EAT. For comparison, we also implement the three scheduling using the common approach described in Section~\ref{sec:integration} as control groups: NSS, SSS, and OSS. The six AMOD  systems share the same experimental setup except the dispatching approaches.

In the simulation, the vehicle speed limit is 25 Miles/hour (40.2336 Km/hour). The maximum load capacity of an autonomous vehicle is 4 and the maximum speed capacity is 100 Miles/hour. The $A^{\ast}$ algorithm is selected as the routing algorithm. All other parameter settings are default as used in the \textit{MobilityTestbed}. The simulation is conducted on a quad-core 2.3 GHz machine with 16 GB of RAM.
\subsection{Experimental Results}
We compare both the average passenger waiting time (see Table~\ref{table:apwt}) and the trip success rate  (see Table~\ref{table:tsr}) of the AMOD systems using the six different dispatching approaches with the 2013 New York City Taxi Data (from 01-01-2013 to 12-31-2013).

Table~\ref{table:apwt} and Table~\ref{table:improve} show that the EAT algorithm significantly reduces the average waiting time, both monthly and yearly, with all the three scheduling strategies. When the AMOD system follows the no-scheduling strategy, the EAT algorithm shortens the monthly average passenger waiting time by up to $49.74\%$ (2.82 $ mins$). The average passenger waiting time of the whole year is reduced by $29.82\%$, from 8.14 $mins$ to 6.27 $mins$. Considering the total number of the trips in the year 2013, it saves 3,870,245.09 hours' time for the passengers as well as substantial reduction on operational cost and green gas emission (though we do not calculate it due to limited availability of the parameters). In the static-scheduling strategy scenarios, the EAT algorithm improves the system's performance by $26.42\%$, bringing the overall average waiting time down from 7.80 $mins$ to 6.17 $mins$. As for the systems with online-scheduling strategy, the yearly average waiting time is also diminished greatly by $26.51\%$.  Though the EAT algorithm considerably improves systems' performance irrespective of the the scheduling strategies, it works best when associated with the NSS scenario. This is because it is more prone to local minimum in the NSS scenario than the others, while the EAT algorithm counteracts the effect by systematically expanding the search space and updating the dispatching adjacency schedule.

 From Table ~\ref{table:tsr} and Table ~\ref{table:improve}, we can see that the EAT algorithms  increases the trip success rate for systems with all the three scheduling strategies. The improvement for both NSS and SSS systems is slight, and below $5\%$ in most months of the year. However, the improvement for the OSS system is significant, and can be up to $11.12\%$ monthly and  $7.65\%$ for the whole year. The reason behind is that the combination of OSS and EAT provides the dispatcher sufficient search space of vehicles and updated information of the traffic information to target an optimal assignment.
 
\begin{figure}
\centering
\caption{Daily average passenger waiting time of AMOD systems using six different dispatching approaches (New York Taxi Data, April 2013)}
\includegraphics[width=0.84\textwidth]{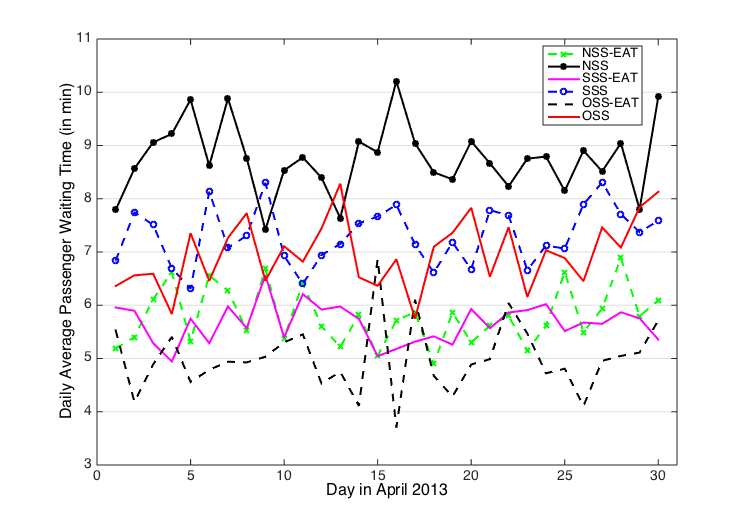}
\label{fig:dapwt}
\end{figure}

\begin{figure}
\centering
\caption{Daily trip success rate (in percentage $\%$) of AMOD systems using six different dispatching approaches (New York Taxi Data, April 2013)}
\includegraphics[width=0.81\textwidth]{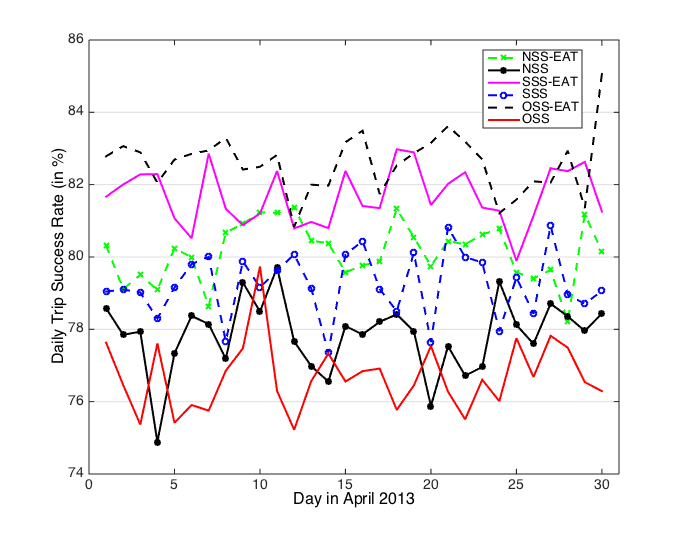}
\label{fig:dtsr}
\end{figure}

 Fig.~\ref{fig:dapwt} and \ref{fig:dtsr} shows the daily average passenger waiting time and the trip success rate of the AMOD systems using the six dispatching paradigms in April 2013. They demonstrate that the OSS-EAT system performs the best among all the six systems, in measurement of both daily average passenger waiting time and trip success rate. The NSS system has the highest daily average waiting time in most case, however,  it performs better than OSS according to the comparison of daily trip success rate. The reason is not clearly known to us yet, though it may be owing to the OSS system's constantly updating of the traffic information.
 
In summary, the EAT algorithm significantly improves the performance of AMOD systems with all the three scheduling strategies according to both the metrics. Specifically, it significantly reduces the average passenger waiting time by up to $29.82\%$. It increases the trip success rate by up to $7.65\%$.

\begin{table*}\centering
\caption{A comparison of the average passenger waiting time (in minutes) of the AMOD systems using the six different dispatching approaches with the 2013 New York City Taxi Data.}
\ra{1.3}
\begin{tabular}{@{}rrrcrrcrrcrrc@{}}\toprule
& \multicolumn{2}{c}{NSS} & \phantom{abc}& \multicolumn{2}{c}{SSS} & \phantom{abc} & \multicolumn{2}{c}{OSS} \\
\cmidrule{2-3} \cmidrule{5-6} \cmidrule{8-9}
& w/ EAT $\quad$  & w/o EAT  && w/ EAT $\quad$  & w/o  EAT  && w/ EAT $\quad$  & w/o EAT  \\ 
\midrule
$Jan \quad$ & $6.71 \quad$ & $ 8.87$ && $6.68\quad$&	$8.01$ &&	$6.03\quad$ &	$7.45$  \\
$Feb \quad$ & $6.99 \quad$ & $9.11$ && $6.82\quad$ &	$8.17$ &&	$6.12\quad$ &	$7.62$ \\
$Mar \quad$ & $7.43 \quad$ & $9.21$ && $7.39\quad$ &	$9.15$ &&	$6.98\quad$ &	$8.51$ \\
$Apr \quad$ & $5.67 \quad$ &	$8.49$ &&	$5.66\quad$ &	$7.23$ &&	$5.01\quad$ &	$6.74$ \\
$May \quad$ & $5.11 \quad$ &	$7.30$ &&	$5.04\quad$ &	$6.78$ &&	$4.78\quad$ &	$6.08$ \\
$June \quad$ & $6.85 \quad$ &    $8.43$ &&   $6.76\quad$ &   $8.10$ &&   $6.14\quad$ &   $7.97$ \\
$July \quad$ & $6.68 \quad$ &    $8.56$ &&   $6.37\quad$ &   $7.81$ &&   $6.19\quad$ &   $7.55$ \\
$Aug \quad$ & $4.88 \quad$ &    $6.67$ &&   $4.85\quad$ &   $6.50$ &&   $4.64\quad$ &   $6.16$ \\
$Sept \quad$ & $8.60 \quad$ &    $10.28$ &&   $8.52\quad$ &   $10.21$ &&   $7.33\quad$ &   $8.68$ \\
$Oct \quad$ & $6.98 \quad$ &    $8.83$ &&   $6.83\quad$ &   $8.75$ &&   $6.17\quad$ &   $7.78$ \\
$Nov \quad$ & $5.89\quad$ &    $7.64$ &&   $5.73\quad$ &   $7.65$ &&   $5.22\quad$ &   $7.06$ \\
$Dec \quad$ & $4.16 \quad$ &    $5.65$ &&   $4.05\quad$ &   $6.18$ &&   $3.91\quad$ &   $5.30$ \\
$year \quad$ & $6.27 \quad$ &   $8.14$&&    $6.17\quad$ &   $7.80$ &&    $5.62\quad$  &  $7.11$ \\
\bottomrule
\end{tabular}
\label{table:apwt}
\end{table*}

\begin{table*}\centering
\caption{A comparison of the trip success rate(in percentage $\%$) of the AMOD systems using the six different dispatching approaches with the 2013 New York City Taxi Data.}
\ra{1.3}
\begin{tabular}{@{}rrrcrrcrrcrrc@{}}\toprule
& \multicolumn{2}{c}{NSS} & \phantom{abc}& \multicolumn{2}{c}{SSS} & \phantom{abc} & \multicolumn{2}{c}{OSS} \\
\cmidrule{2-3} \cmidrule{5-6} \cmidrule{8-9}
& w/ EAT $\quad$  & w/o EAT  && w/ EAT $\quad$  & w/o  EAT  && w/ EAT $\quad$  & w/o EAT  \\ 
\midrule
$Jan \quad$ & $87.32 \quad$ & $ 79.96$ && $89.44\quad$&	$87.46$ &&	$90.79\quad$ &	$86.87$  \\
$Feb \quad$ & $83.67 \quad$ & $89.04$ && $92.69\quad$ &	$89.61$ &&	$93.63\quad$ &	$89.64$ \\
$Mar \quad$ & $92.16 \quad$ & $89.04$ && $92.69\quad$ &	$89.61$ &&	$93.63\quad$ &	$89.64$ \\
$Apr \quad$ & $80.20 \quad$ &	$77.70$ &&	$81.93\quad$ &	$79.26$ &&	$82.31\quad$ &	$76.82$ \\
$May \quad$ & $89.96 \quad$ &	$88.10$ &&	$94.05\quad$ &	$89.69$ &&	$96.77\quad$ &	$90.96$ \\
$June \quad$ & $83.11 \quad$ &    $80.66$ &&   $85.34\quad$ &   $83.07$ &&   $89.27\quad$ &   $81.81$ \\
$July \quad$ & $86.05 \quad$ &    $84.76$ &&   $89.14\quad$ &   $85.17$ &&   $91.79\quad$ &   $85.61$ \\
$Aug \quad$ & $89.27 \quad$ &    $84.79$ &&   $91.09\quad$ &   $87.86$ &&   $94.01\quad$ &   $85.71$ \\
$Sept \quad$ & $79.87 \quad$ &    $76.04$ &&   $82.31\quad$ &   $80.00$ &&   $85.91\quad$ &   $80.48$ \\
$Oct \quad$ & $82.79 \quad$ &    $80.55$ &&   $85.32\quad$ &   $81.62$ &&   $89.79\quad$ &   $81.35$ \\
$Nov \quad$ & $74.99\quad$ &    $72.61$ &&   $78.35\quad$ &   $71.65$ &&   $81.27\quad$ &   $76.51$ \\
$Dec \quad$ & $79.52 \quad$ &   $76.23$&&    $83.65\quad$ &   $81.40$ &&    $87.66\quad$  &  $78.89$ \\
$year \quad$ & $83.91 \quad$ &    $80.97$ &&   $86.79\quad$ &   $83.12$ &&   $89.59\quad$ &   $83.22$ \\
\bottomrule
\end{tabular}
\label{table:tsr}
\end{table*}

\begin{table*}\centering
\caption{Performance improvement (in percentage $\%$) of the AMOD systems on average passenger waiting time ($T_{APW}$ $(min)$) and trip success rate ($R_{TS}$ $\%$) brought by the  \textit{Expand and Target} algorithm.}
\ra{1.3}
\begin{tabular}{@{}rrrcrrcrrcrrc@{}}\toprule
& \multicolumn{2}{c}{NSS} & \phantom{abc}& \multicolumn{2}{c}{SSS} & \phantom{abc} & \multicolumn{2}{c}{OSS} \\
\cmidrule{2-3} \cmidrule{5-6} \cmidrule{8-9}
& $T_{APW }(min)$   & $R_{TS} (\%)$  && $T_{APW}(min)$   & $R_{TS}(\%)$  && $T_{APW}(min)$   & $R_{TS}(\%)$  \\ 
\midrule

$Jan \quad$ & $32.19 \quad$ & $ \textbf{9.20}$ && $19.91\quad$&	$2.26$ &&	$23.55\quad$ &	$4.51$  \\
$Feb \quad$ & $30.33 \quad$ & $1.63$ && $19.79\quad$ &	$5.17$ &&	$24.51\quad$ &	$6.29$ \\
$Mar \quad$ & $23.96 \quad$ & $3.50$ && $23.82\quad$ &	$3.44$ &&	$21.92\quad$ &	$4.45$ \\
$Apr \quad$ & $\textbf{49.74} \quad$ &	$3.22$ &&	$27.74\quad$ &	$3.37$ &&	$34.53\quad$ &	$7.15$ \\
$May \quad$ & $42.86 \quad$ &	$2.11$ &&	$34.52\quad$ &	$4.86$ &&	$27.20\quad$ &	$6.39$ \\
$June \quad$ & $23.07 \quad$ &    $3.04$ &&   $19.82\quad$ &   $2.73$ &&   $29.80\quad$ &   $9.12$ \\
$July \quad$ & $28.14 \quad$ &    $1.52$ &&   $22.61\quad$ &   $4.66$ &&   $21.97\quad$ &   $7.22$ \\
$Aug \quad$ & $36.68 \quad$ &    $5.28$ &&   $34.02\quad$ &   $3.68$ &&   $32.76\quad$ &   $9.68$ \\
$Sept \quad$ & $19.53 \quad$ &    $5.04$ &&   $19.84\quad$ &   $2.89$ &&   $18.42\quad$ &   $6.75$ \\
$Oct \quad$ & $26.50 \quad$ &    $2.78$ &&   $28.11\quad$ &   $4.53$ &&   $26.09\quad$ &   $10.37$ \\
$Nov \quad$ & $29.71\quad$ &    $3.28$ &&   $31.94\quad$ &   $\textbf{9.35}$ &&   $35.25\quad$ &   $6.22$ \\
$Dec \quad$ & $35.82 \quad$ &    $4.32$ &&   $\textbf{52.59}\quad$ &   $2.76$ &&   $\textbf{35.55}\quad$ &   $\textbf{11.12}$ \\
$year \quad$ & $29.82 \quad$ &   $3.63$&&    $26.42\quad$ &   $4.42$ &&    $26.51\quad$  &  $7.65$ \\
\bottomrule
\end{tabular}
\label{table:improve}
\end{table*}


\section{Conclusion and Future Work}
\label{sec:conclusion}
 Autonomous mobility on demand systems, which, though still in their infancy, have very promising prospects in providing urban population with sustainable and safe personal mobility service in the near future. While a lot of research has been done on autonomous vehicles and mobility on demand systems, to the best of our knowledge, this is the first work that shows how to manage autonomous mobility on demand systems for better passenger experience.
 
To reduce the average passenger waiting time and increase the trip success rate, we introduce the \textit{Expand and Target} algorithm which can be easily integrated with three different scheduling strategies for dispatching autonomous vehicles. We implement the autonomous mobility on demand simulation platform and conduct an empirical study with the 2013 New York City Taxi Data. Experimental results demonstrate that the algorithm significantly improves the performance of the autonomous mobility on demand systems: it reduces the overall average passenger waiting time by up to $29.82\%$ and increases the trip success rate by up to $7.65\%$; it saves millions of hours of passengers' time annually.

While the results are impressive, there is still a long way to go towards a society with near instantly available personal mobility. To facilitate research in this emerging field, we are developing a robust, open-source, agent-based simulation platform for autonomous mobility on demand systems from scratch.  Another interesting direction is to investigate and design novel mechanisms to encourage passengers truthfully report their travel demand  well in advance to the dispatcher so that better performance can be achieved. Moreover, it would be useful to  study machine learning techniques that can accurately predict passengers' mobility patterns based on historical data.

\section*{Acknowledgments.} The authors would like to thank Michal Jakob, Michal Certicky and Martin Schaefer for sharing the source code of the \textit{MobilityTestbed}  and  the \textit{AgentPolis} platform, and providing other technical support regarding the development of the AMOD simulation tool.


\begin{thebibliography}{10}
\providecommand{\url}[1]{\texttt{#1}}
\providecommand{\urlprefix}{URL }

\bibitem{agussurja2012toward}
Agussurja, L., Lau, H.C.: Toward large-scale agent guidance in an urban taxi
  service. arXiv preprint arXiv:1210.4849  (2012)

\bibitem{alshamsi2009multiagent}
Alshamsi, A., Abdallah, S., Rahwan, I.: Multiagent self-organization for a taxi
  dispatch system. In: 8th International Conference on Autonomous Agents and
  Multiagent Systems. pp. 21--28 (2009)

\bibitem{anderson2014autonomous}
Anderson, J.M., Nidhi, K., Stanley, K.D., Sorensen, P., Samaras, C., Oluwatola,
  O.A.: Autonomous vehicle technology: A guide for policymakers. Rand
  Corporation (2014)

\bibitem{beirao2007understanding}
Beir{\~a}o, G., Cabral, J.S.: Understanding attitudes towards public transport
  and private car: A qualitative study. Transport policy  14(6),  478--489
  (2007)

\bibitem{broggi2013extensive}
Broggi, A., Buzzoni, M., Debattisti, S., Grisleri, P., Laghi, M.C., Medici, P.,
  Versari, P.: Extensive tests of autonomous driving technologies. Intelligent
  Transportation Systems, IEEE Transactions on  14(3),  1403--1415 (2013)

\bibitem{vcerticky2015analyzing}
{\v{C}}ertick{\`y}, M., Jakob, M., P{\'\i}bil, R.: Analyzing on-demand mobility
  services by agent-based simulation. Journal of Ubiquitous Systems \&
  Pervasive Networks  6(1),  17--26 (2015)

\bibitem{vcerticky2014agent}
{\v{C}}ertick{\`y}, M., Jakob, M., P{\'\i}bil, R., Moler, Z.: Agent-based
  simulation testbed for on-demand transport services. In: Proceedings of the
  2014 international conference on Autonomous agents and multi-agent systems.
  pp. 1671--1672 (2014)

\bibitem{chong2013autonomy}
Chong, Z., Qin, B., Bandyopadhyay, T., Wongpiromsarn, T., Rebsamen, B., Dai,
  P., Rankin, E., Ang~Jr, M.H.: Autonomy for mobility on demand. In:
  Intelligent Autonomous Systems 12, pp. 671--682. Springer (2013)

\bibitem{delling2009engineering}
Delling, D., Sanders, P., Schultes, D., Wagner, D.: Engineering route planning
  algorithms. In: Algorithmics of large and complex networks, pp. 117--139.
  Springer (2009)

\bibitem{dan2014newyork}
Donovan, B., Work, D.: New york city taxi data 2010–2013.
  http://publish.illinois.edu/dbwork/open-data/  (2014)

\bibitem{downs2005still}
Downs, A.: Still stuck in traffic: coping with peak-hour traffic congestion.
  Brookings Institution Press (2005)

\bibitem{gan2015optimizing}
Gan, J., An, B., Miao, C.: Optimizing efficiency of taxi systems: Scaling-up
  and handling arbitrary constraints. In: Proceedings of the 2015 International
  Conference on Autonomous Agents and Multiagent Systems. pp. 523--531 (2015)

\bibitem{glaschenko2009multi}
Glaschenko, A., Ivaschenko, A., Rzevski, G., Skobelev, P.: Multi-agent real
  time scheduling system for taxi companies. In: 8th International Conference
  on Autonomous Agents and Multiagent Systems (AAMAS 2009), Budapest, Hungary.
  pp. 29--36 (2009)

\bibitem{haklay2008openstreetmap}
Haklay, M., Weber, P.: Openstreetmap: User-generated street maps. Pervasive
  Computing, IEEE  7(4),  12--18 (2008)

\bibitem{huang2009finding}
Huang, A.S., Moore, D., Antone, M., Olson, E., Teller, S.: Finding multiple
  lanes in urban road networks with vision and lidar. Autonomous Robots
  26(2-3),  103--122 (2009)

\bibitem{jakob2012agentpolis}
Jakob, M., Moler, Z., Komenda, A., Yin, Z., Jiang, A.X., Johnson, M.P.,
  P{\v{e}}chou{\v{c}}ek, M., Tambe, M.: Agentpolis: towards a platform for
  fully agent-based modeling of multi-modal transportation. In: Proceedings of
  the 11th International Conference on Autonomous Agents and Multiagent
  Systems-Volume 3. pp. 1501--1502 (2012)

\bibitem{lavrinc2014bmw}
Lavrinc, D.: Bmw builds a self-driving car - that drifts. Wired Magazine
  (2014)

\bibitem{lozano2012autonomous}
Lozano-Perez, T., Cox, I.J., Wilfong, G.T.: Autonomous robot vehicles. Springer
  Science \& Business Media (2012)

\bibitem{maciejewski2013simulation}
Maciejewski, M., Nagel, K.: Simulation and dynamic optimization of taxi
  services in matsim. VSP Working Paper  (2013)

\bibitem{mapzen2015nyc}
MAPZEN: Mapzen metro extracts: New york city.
  https://mapzen.com/data/metro-extracts  (2015)

\bibitem{markoff2010google}
Markoff, J.: Google cars drive themselves, in traffic. New York Times  (2010)

\bibitem{mitchell2007intelligent}
Mitchell, W.J.: Intelligent cities. UOC papers  5,  1885--1541 (2007)

\bibitem{mitchell2010reinventing}
Mitchell, W.J.: Reinventing the automobile: Personal urban mobility for the
  21st century. MIT Press (2010)

\bibitem{moghadam2008improving}
Moghadam, P., Wijesoma, W.S., Feng, D.J.: Improving path planning and mapping
  based on stereo vision and lidar. In: Proceedings of the 10th International
  Conference on Control, Automation, Robotics and Vision. pp. 384--389. IEEE
  (2008)

\bibitem{pediacities2015nyc}
Ontodia: Pediacities neighborhoods of new york city.
  http://catalog.opendata.city/dataset/pediacities-nyc-neighborhoods  (2015)

\bibitem{papadimitratos2009vehicular}
Papadimitratos, P., La~Fortelle, A., Evenssen, K., Brignolo, R., Cosenza, S.:
  Vehicular communication systems: Enabling technologies, applications, and
  future outlook on intelligent transportation. Communications Magazine, IEEE
  47(11),  84--95 (2009)

\bibitem{premebida2009lidar}
Premebida, C., Ludwig, O., Nunes, U.: Lidar and vision-based pedestrian
  detection system. Journal of Field Robotics  26(9),  696--711 (2009)

\bibitem{seow2010collaborative}
Seow, K.T., Dang, N.H., Lee, D.H.: A collaborative multiagent taxi-dispatch
  system. Automation Science and Engineering, IEEE Transactions on  7(3),
  607--616 (2010)

\bibitem{spieser2014toward}
Spieser, K., Treleaven, K., Zhang, R., Frazzoli, E., Morton, D., Pavone, M.:
  Toward a systematic approach to the design and evaluation of automated
  mobility-on-demand systems: A case study in singapore. In: Road Vehicle
  Automation, pp. 229--245. Springer (2014)

\bibitem{zhang2014control}
Zhang, R., Pavone, M.: Control of robotic mobility-on-demand systems: a
  queueing-theoretical perspective. arXiv preprint arXiv:1404.4391  (2014)

\end{thebibliography}
\end{document}